\def\year{2022}\relax
\documentclass[letterpaper]{article} 
\usepackage{aaai22}  
\usepackage{times}  
\usepackage{helvet}  
\usepackage{courier}  
\usepackage[hyphens]{url}  
\usepackage{graphicx} 
\usepackage{natbib}  
\usepackage{caption} 
\usepackage{algorithm}
\usepackage{algorithmic}

\usepackage{multirow}
\usepackage{booktabs}
\usepackage{makecell}

\title{LMR-CBT: Learning Modality-fused Representations with CB-Transformer for Multimodal Emotion Recognition from Unaligned Multimodal Sequences}

\author{
    Ziwang Fu\textsuperscript{\rm 1},
    Feng Liu\textsuperscript{\rm 2},
    Hanyang Wang\textsuperscript{\rm 2}, 
    Siyuan Shen\textsuperscript{\rm 2},
    Jiahao Zhang\textsuperscript{\rm 2},
    Jiayin Qi\textsuperscript{\rm 3},
    Xiangling Fu\textsuperscript{\rm 1},
    Aimin Zhou\textsuperscript{\rm 2}
    \\
}
\affiliations{
    \textsuperscript{\rm 1} Beijing University of Posts and Telecommunications, Beijing, China\\
    \textsuperscript{\rm 2} East China Normal University, Shanghai, China\\
    \textsuperscript{\rm 3} Shanghai University of International Business and Economics, Shanghai, China\\
    Email: fzw150355@163.com, lsttoy@163.com, faceeyes@126.com, trert9811@gmail.com, 10185102243@stu.ecnu.edu.cn, qijiayin@139.com, fuxiangling@bupt.edu.cn, amzhou@cs.ecnu.edu.cn

}

\usepackage{bibentry}

\begin{document}

\maketitle

\begin{abstract}
Learning modality-fused representations and processing unaligned multimodal sequences are meaningful and challenging in multimodal emotion recognition. Existing approaches use directional pairwise attention or a message hub to fuse language, visual, and audio modalities. However, those approaches introduce information redundancy when fusing features and are inefficient without considering the complementarity of modalities. In this paper, we propose an efficient neural network to learn modality-fused representations with CB-Transformer (LMR-CBT) for multimodal emotion recognition from unaligned multimodal sequences. Specifically, we first perform feature extraction for the three modalities respectively to obtain the local structure of the sequences. Then, we design a novel transformer with cross-modal blocks (CB-Transformer) that enables complementary learning of different modalities, mainly divided into local temporal learning, cross-modal feature fusion and global self-attention representations. In addition, we splice the fused features with the original features to classify the emotions of the sequences. Finally, we conduct word-aligned and unaligned experiments on three challenging datasets, IEMOCAP, CMU-MOSI, and CMU-MOSEI. The experimental results show the superiority and efficiency of our proposed method in both settings. Compared with the mainstream methods, our approach reaches the state-of-the-art with a minimum number of parameters.
\end{abstract}

\section{Introduction}

Multimodal emotion recognition has attracted increasing attention due to its robustness and remarkable performance \cite{articletmp, articletrans, dai2021multimodal}. The goal of this task is to recognize human emotions from video clips, which involves three main modalities: natural language, facial expressions and audio signals. Emotion recognition is applied in areas such as social robotics, educational quality assessment, and healthcare, where the analysis of emotion is particularly important during COVID-19 \cite{chandra2021covid19}. Multimodality provides a wealth of information compared to single modality and can fully reflect emotional states. However, due to the different sampling rates of sequences from different modalities, the collected multimodal states are often unaligned. Manually aligning different modalities is often labor-intensive and requires domain knowledge \cite{tsai2019learning,mctn}. In addition, most of the networks with high performance cannot achieve a balance between the number of parameters and performance. To this end, we focus on the ability to learn the representation of fused modalities and efficiently perform multimodal emotion recognition on unaligned sequences.

\begin{figure}
    \centering
   \includegraphics[width=\linewidth]{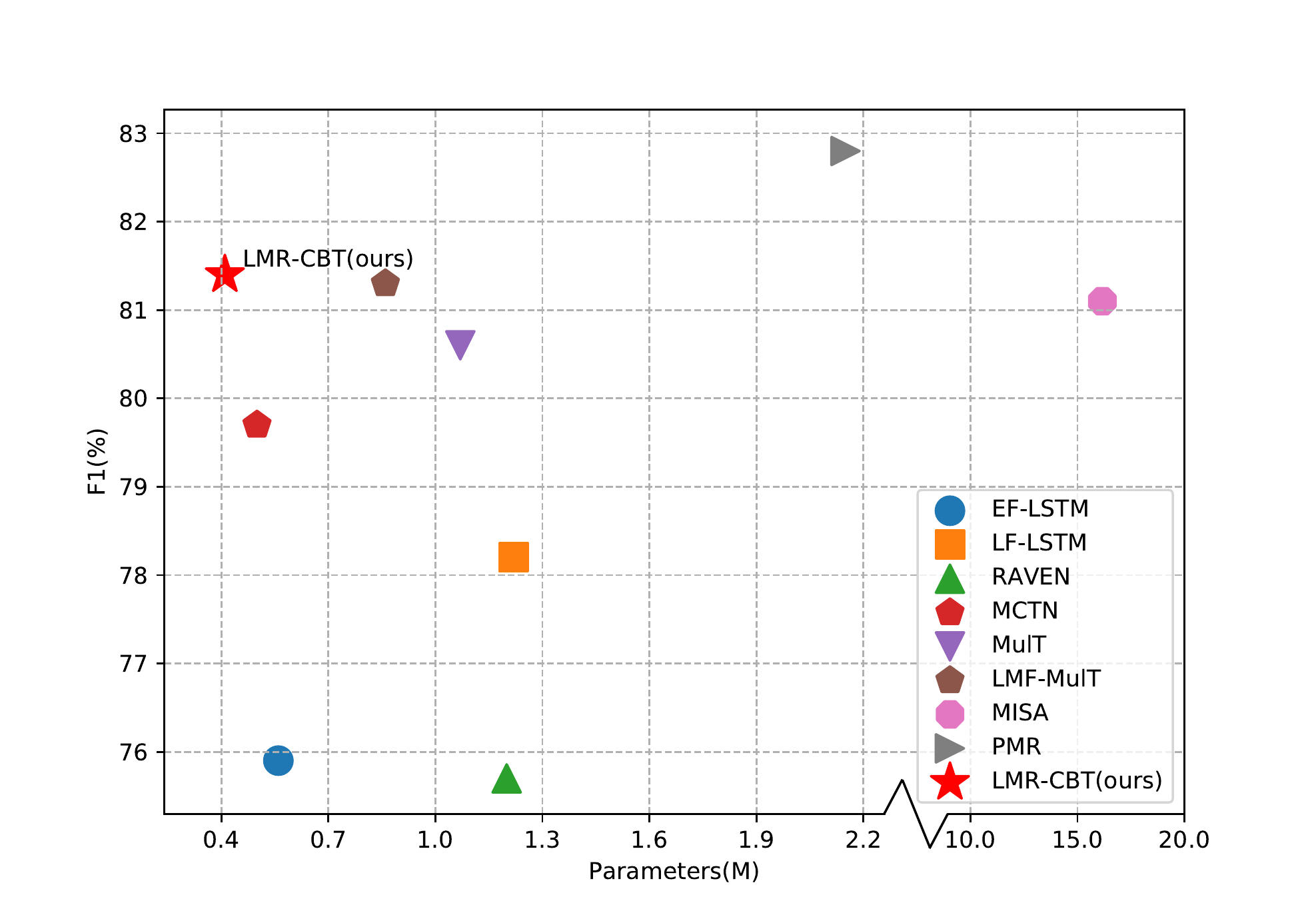}
   \caption{Comparisons of different methods on CMU-MOSEI in terms of F1 score and the number of parameters. The proposed LMR-CBT model achieves the best performance with an order of magnitude smaller model size.}
   \vspace{-10pt}
   \label{fig:params}
\end{figure}

\begin{figure*}[!t]
\centering
\vspace{-0.5em}
\includegraphics[scale =0.6]{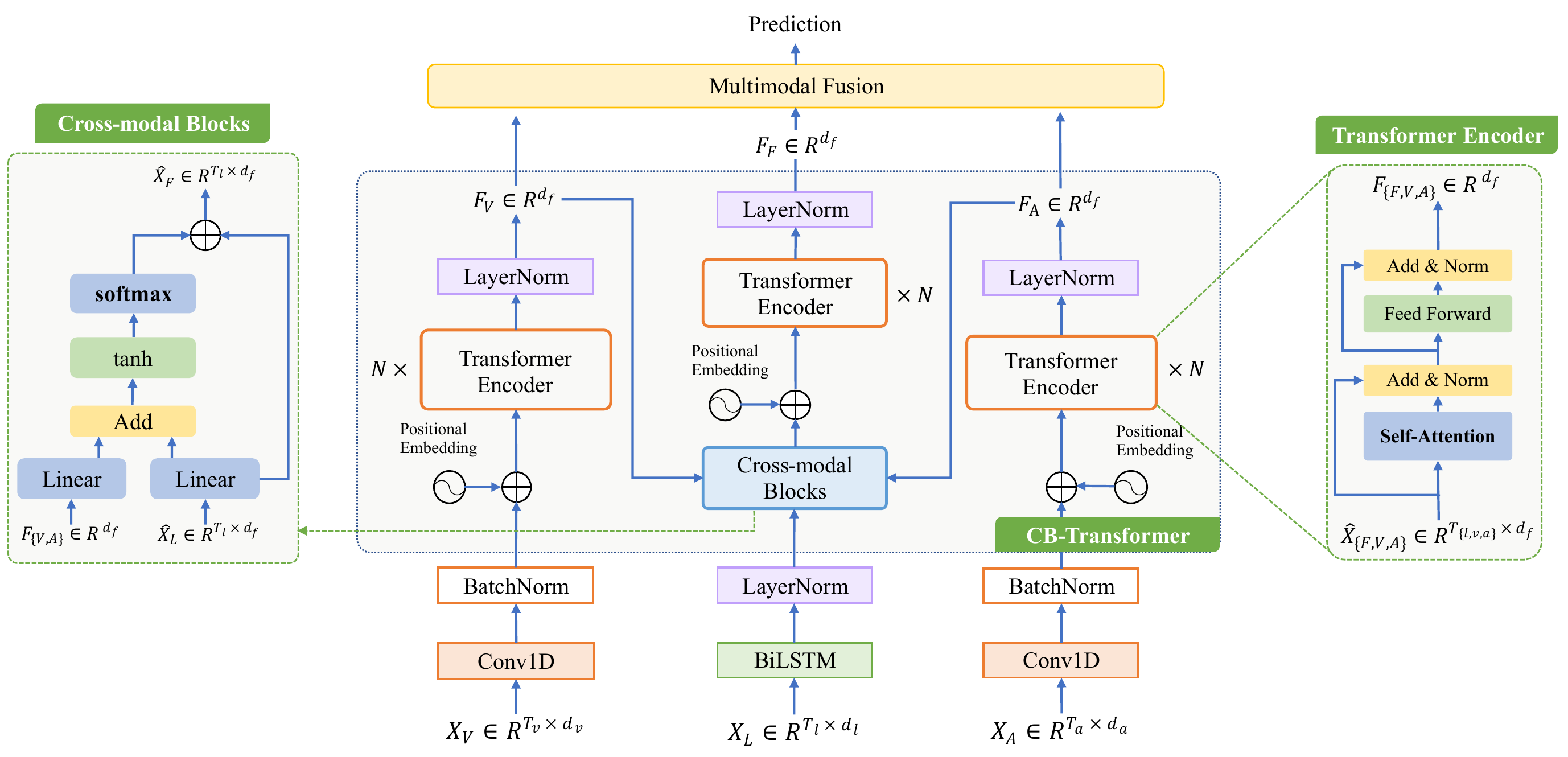}
\caption{The overall architecture of LMR-CBT. \textbf{Middle:} we design a novel transformer with cross-modal blocks (CB-Transformer) that can make different modals complementary learning, which is mainly divided into local temporal learning, cross-modal feature fusion and global self-attention representations. \textbf{Left:} the structure of the residual-based cross-modal fusion method. \textbf{Right:} the structure of the transformer encoder.}
\vspace{-0.5em}
\label{fig:framework}
\end{figure*}

In the previous works \cite{sahay2020low, rahman2020integrating, hazarika2020misa, yu2021learning, dai2021weaklysupervised}, Transformers \cite{10.5555/3295222.3295349} are mostly used for unaligned multimodal emotion recognition. Typically, \citet{tsai2019multimodal} proposed the Multimodal Transformer (MulT) method to fuse information from different modalities in unaligned sequences without explicitly aligning the data. The approach learns the interactions between pairs of elements through a cross-modal attention module that iteratively reinforces features of one modality with features of other modalities. Recently, \citet{Lv_2021_CVPR} proposed the Progressive Modality Reinforcement (PMR) by introducing a message hub to exchange information with each modality. The approach uses a progressive strategy to utilize high-level source modality information for unaligned multimodal sequences fusion.

However, MulT only considers the fusion of features between modality pairs, ignoring the coordination of the three modalities. Besides, using a pairwise approach to fuse the modal features can produce redundant information. For example, the visual representations are repeated twice in the concatenation of  visual-language features and visual-audio features. PMR considers the association among the three modalities, but fusing the modal features by designing a centralized message hub would sacrifice its efficiency. To be more specific, the information of the three modalities needs to interact closely and recursively with the message hub to ensure the integrity of the features, and such an operation requires a huge number of parameters. Meanwhile, this approach does not take into account the complementarity between modal information, while feature fusion can be accomplished by simply using the interaction between modalities without introducing a third party. What's more, recent methods are too high in the number of parameters to be applicable to realistic scenarios due to pre-trained models.

Therefore, to address the above limitations, we propose a neural network to learn modality-fused representations with CB-Transformer (LMR-CBT) for multimodal emotion recognition from unaligned multimodal sequences. Figure \ref{fig:framework} shows the overall architecture of LMR-CBT. Specifically, we first perform feature extraction for the three modalities respectively to obtain the local structure of the sequences. For the audio and visual modalities, we obtain information about adjacent elements by 1D temporal convolution. For the language modality, we use Bi-directional Long and Short Term Memory (BiLSTM) to capture the long term dependencies and the contextual information between texts.

After obtaining feature representations of the three modalities, we design a novel transformer with cross-modal blocks (CB-Transformer) to achieve complementary learning of the different modalities, which is mainly divided into local temporal learning, cross-modal feature fusion and global self-attention representations. In the local temporal learning part, audio and visual features are used to obtain adjacent element-dependent representations of the two modalities through the transformer. In the cross-modal feature fusion part, residual-based modal interaction approach is used to obtain the fused features of the three modalities. In the global self-attention representations part, the transformer learns high-level representations within the fusion modality. The CB-Transformer can adequately represent the fused features without losing the original features and can efficiently handle unaligned multimodal sequences. Finally, we splice the modal fusion features with the original features to obtain the emotional categories. We perform word-aligned and unaligned experiments on three mainstream public datasets of multimodal emotion recognition, IEMOCAP \cite{iemocap}, CMU-MOSI \cite{cmumosi} and CMU-MOSEI \cite{cmu-mosei}. The experimental results demonstrate the superiority of our proposed method. Moreover, we achieve a better trade-off between the performance and the efficiency. Compared with the mainstream methods, our approach reaches the state-of-the-art with a minimum number of parameters.

We summarize our three main contributions as follows:

\begin{itemize}
    \item We propose an efficient neural network to learn modality-fused representations with CB-Transformer (LMR-CBT) for multimodal emotion recognition from unaligned multimodal sequences (only 0.41M), which can effectively fuse the interactive information of the three modalities.
    \item We design a novel transformer with cross-modal blocks (CB-Transformer) to achieve complementary learning of different modalities, which is mainly divided into local temporal learning, cross-modal feature fusion and global self-attention representations. The CB-Transformer can adequately represent the fused features without losing the original features and can efficiently handle unaligned multimodal sequences.
    \item We obtain a better trade-off between the performance and the efficiency on three challenging datasets. Compared with the existing state-of-the-art methods, LMR-CBT achieves comparable or even higher performance with a minimal number of parameters.
\end{itemize}

\section{Related Work}

Multimodal emotion recognition has attracted a lot of attention in recent years. This task requires the fusion of cross-modal information of temporal sequential signals. According to the approaches of feature fusion, it can be divided into early fusion \cite{10.1145/2070481.2070509,inproceedings2013early}, late fusion \cite{article2016late, 80193012017late} and model fusion. Previous works have focused on early or late fusion strategies. Early fusion strategies involve fusing the shallow inter-modal features and focusing on mixed-modal feature processing while late fusion strategies involve finding the confidence level of each modality and then coordinating them to make joint decisions. Although better performance can be obtained using these fusion strategies in comparison to single modality learning, they do not explicitly consider the intrinsic connection between sequence elements from different modalities, which is essential for effective multimodal fusion. Subsequently, model fusion is gradually applied and more complicated models are proposed. \citet{2019work} used visual and auditory features to shift words in text with attention. \citet{rahman2020integrating} introduced a multimodal adaptive gate that integrates visual and acoustic information into a large pretrained language model. \citet{hazarika2020misa} incorporated a combination of losses including distributional similarity, orthogonal loss, reconstruction loss and task prediction loss to learn modality-invariant and modality-specific representation. \citet{dai2021multimodal} introduced sparse cross-attention to achieve end-to-end emotion recognition. \citet{dai2021weaklysupervised} proposed a multi-task learning approach using weak supervision for multimodal emotion recognition. \citet{yu2021learning} proposed a way to fuse features from different modalities by combining self-supervised and multi-task learning. Although self-supervised and multi-task learning can effectively alleviate the problem of small samples, how to perform efficient cross-modal interactions is still a tremendously challenging issue for researchers. Therefore, the main motivation of this work is how to perform unaligned multimodal emotion recognition with a minimalist design excluding tricks like self-supervision or multi-tasking.

In order to fuse the information of unaligned multimodal sequences, early works have explored the dependencies between modal elements based on the maximum modal information criterion \cite{1467547}. However, the performance of those early approaches is far from satisfactory due to the shallow model structures. \citet{tsai2019multimodal} proposed a multimodal transformer (MulT) to learn inter-modal correlations using a cross-modal attention mechanism. \citet{sahay2020low} proposed low rank fusion based transformers (LMT-MULT) to design LMF units for efficient modal feature fusion based on previous work. \citet{Lv_2021_CVPR} proposed progressive modality reinforcement (PMR) method. This method uses a message hub to interact with the three modal information and adopts the progressive strategy to fuse unaligned multimodal temporal sequences utilizing high level source modal information. Although those previous trials have made some performance improvement in unaligned multimodal emotion recognition, they still faces the problems of effective fusion of cross-modal features and the inability to ensure that information is not lost. In this paper, we mainly focus on reaching an accuracy-parameter balance by a novel information redundancy-free modal fusion strategy.

\section{Methodology}

\subsection{Problem statement}

The multimodal emotion recognition task mainly involves three modalities, language$(L)$, visual$(V)$ and audio$(A)$. We define that the three modalities are obtained through feature extraction as $ X_{\{L,V,A\}} \in {R}^{T_{\{l,v,a\}} \times d_{\{l,v,a\}}} $, where $T_{(.)}$ represents the length of the sequence and $d_{(.)}$ represents dimensions of the extracted features. Our goal is to efficiently extract features of different modalities from the unaligned multimodal sequences and to obtain a fused representation across modalities. We expect the multimodal representation to accurately predict the emotion category of the sequence.

\subsection{Overall Architecture}

We propose a neural network to learn modality-fused representations with CB-Transformer (LMR-CBT), and the overall architecture of the network is shown in Figure \ref{fig:framework}, where the cross-modal blocks and transformer encoder in CB-Transformer are located on the two sides of the figure, respectively. Next, we will describe the network in detail.

\subsubsection{Feature Preprocessing.}

The feature preprocessing is performed separately according to the temporal structure of the different modalities. For audio and visual modalities, to ensure that each element in the input sequence has sufficient perception of its neighboring elements, we put the two modalities into 1D temporal convolution separately by setting different convolution kernel sizes. The specific formula is as follows:

\begin{equation}
    \hat{X}_{\{V,A\}} = BN(Conv1D(X_{\{V,A\}}, k_{\{V,A\}})) \in {R}^{T_{\{v,a\}} \times d_f}
\end{equation}
where $BN$ stands for batch normalization, and $k_{\{V,A\}}$ is the size of the convolution kernel of modality $\{V,A\}$, and $d_f$ represents a common dimension. 

In terms of language modality, we consider that the language itself is characterized by long-time dependencies and associative contextual information. BiLSTM can better capture bidirectional long-time semantic dependencies and identify the emotional representations of languages. We use a two-layer BiLSTM for feature extraction:

\begin{equation}
    \hat{X}_L = LN(BiLSTM(X_L)) \in {R}^{T_l \times d_f}
\end{equation}
where $LN$ represents layer normalization. The purpose of layer normalization is to stabilize the distribution of each layer so that subsequent layers can learn the content of the previous layer in a stable manner. By the above operation, on the one hand, we can aggregate the features of adjacent elements, and on the other hand, we can pre-align the feature dimensions of unaligned multimodal data to the same dimension.

\subsubsection{Transformer with Cross-modal Blocks.}

We design a novel transformer with cross-modal blocks (CB-Transformer). CB-Transformer is divided into three parts: local temporal learning, cross-modal feature fusion and global self-attention representations. In this module, there are two important components: the transformer encoder and the residual-based cross-modal fusion, represented using $TransEncoder$ and $CrossModal$, respectively. For both components we will discuss in detail in Section \ref{sec: te} and \ref{sec: cs}.

In the local temporal learning, we use the transformer encoder, which is becoming increasingly popular in many areas such as computer vision and natural language processing due to its noticeable performance. We use this component to obtain temporal representations of audio and visual modality features that have undergone 1D temporal convolution. The specific process can be expressed by the following formula:

\begin{equation}
    Z^{\left[ 0 \right]}_{\{V,A\}} = \hat{X}_{\{V,A\}} + PE(T_{\{v,a\}}, d_f)
\end{equation}

\begin{equation}
    F_{\{V,A\}} = LN(TransEncoder(Z^{\left[ 0 \right]}_{\{V,A\}})) \in {R}^{d_f}
\end{equation}
where $PE(T_{\{v,a\}}, d_f) \in {R}^{T_{\{v,a\}} \times d_f}$ computes the embeddings for each position index, and $Z^{\left[ 0 \right]}_{\{V,A\}}$ represents the result embedded through position, and $TransEncoder$ represents the transformer encoder, which we will discuss in detail in Section \ref{sec: te}. We use $F_{\{V,A\}}$ to represent the result of local temporal learning.

In the part of cross-modal feature fusion, we design a residual-based cross-modal fusion method, which takes $F_{\{V,A\}}$ and $\hat{X}_L$ as inputs and the fused representation of the three modalities as outputs. The structure of residual can ensure that information is not lost. The specific formula is as follows:

\begin{equation}
    \hat{X}_F = CrossModal(F_{\{V,A\}}, \hat{X}_L) + \hat{X}_L \in {R}^{T_l \times d_f}
\end{equation}
where $CrossModal$ represents the residual-based cross-modal fusion, which we will discuss in detail in Section \ref{sec: cs}, and $\hat{X}_F$ denotes the fusion features. We believe that the fused modal representation not only carries information from the language modality, but also fuses information from all the three modalities to ensure effective interaction of information. Similarly, Transformer Encoder is used to extract the representation of the fused features in the global self-attention representations.

Through global self-attention representations, we can obtain high-level complementary representations of the fused modalities. The specific formula is as follows:

\begin{equation}
    F_{F} =LN(TransEncoder(\hat{X}_{F} + PE(T_{l}, d_f))) \in {R}^{d_f}
\end{equation}
where $F_{F}$ represents the global self-attention learning results for fused representations.

\subsubsection{Prediction.}

We carry out the emotion category prediction. Specifically, we perform a splicing operation on the fused modal representation and the audio/visual original modal representation to obtain $I = [F_{F}, F_{A}, F_{V}]$. After that, we get the final output of the emotional category through the two-layer fully connected network:

\begin{equation}
    prediction = W_2(\sigma(W_1I + b_1)) + b_2 \in {R}^{d_{out}}
\end{equation}
where $d_{out}$ is the output dimensions of emotional categories, $ W_1 \in {R}^{d_f*3}$ and $ W_1 \in {R}^{d_{out}}$ are weight vectors, $b_1$ and $b_2$ are the bias, $\sigma$ denotes the ReLU activation function.

\begin{algorithm}[tb]
\caption{The algorithm of cross-modal fusion}
\label{alg:algorithm}
\textbf{Input}: the audio and video modal representation with local temporal learning $F_{\{V,A\}} \in {R}^{d_f}$; the language representation with BiLSTM processing: $\hat{X}_L \in {R}^{T_l \times d_f}$; the batch size $bs$.\\
\textbf{Output}: the features that fuse the representation of the three modalities and the original text representation: $\hat{X}_F \in {R}^{T_l \times d_f}.$
\begin{algorithmic}[1] 
\STATE $F_{\{V,A\}}$ = Linear($F_{\{V,A\}}$);
\STATE Let attn\_softmax = $\left[ \right]$;
\STATE Let $i$ = 0;
\WHILE{$i < bs$}
\STATE $\hat{X^{*}_L}$ = Linear($\hat{X}_L[i]$);
\STATE $\hat{X^{*}_L}$ += $F_{\{V,A\}}[i]$;
\STATE $\hat{X^{*}_F}$ = softmax(tanh($\hat{X^{*}_L}$));
\STATE attn\_softmax.append($\hat{X^{*}_F}$);
\ENDWHILE
\STATE $\hat{X}_F$ = Concat(attn\_softmax).
\STATE \textbf{return} $\hat{X}_F$.
\end{algorithmic}
\end{algorithm}

\subsection{Transformer Encoder}
\label{sec: te}

We'll introduce the details of transformer encoder used for both local temporal learning and global self-attention representations, as shown on the right side of Figure \ref{fig:framework}. Firstly, following \cite{10.5555/3295222.3295349}, we abstract the data of the temporal series using the sinusoidal position embedding (PE). We encode the positional information of a sequence of length $T$ via the $sin$ and $cos$ functions with frequencies dictated by the feature indices:

\begin{equation}
    Z^{\left[ 0 \right]}_{\{F,V,A\}} = \hat{X}_{\{F,V,A\}} + PE(T_{\{l,v,a\}}, d_f)
\end{equation}

Next, transformer encoder is mainly composed of self-attention, Feedforward and Add\&Norm. Self-Attention is the focus of transformer encoder. The specific formula is as follows:

\begin{equation}
    self\raisebox{0mm}{-}attention(Q,K,V) = softmax(\frac{QK^T}{\sqrt{d_k}})V
\end{equation}
where $Q, K, V$ denotes $Z^{\left[ i-1 \right]}_{\{F,V,A\}}$. $Z^{\left[ i-1 \right]}_{\{F,V,A\}}$ is represented by different projection spaces with different parameter matrices, where $i$ represents the number of layers of transformer attention, $i=1,...,D$.

The Feedforward layer is a two-layer fully connected layer and the activation function of the first layer is Relu:

\begin{equation}
    Z^{\left[ i \right]}_{\{F,V,A\}} = LN( Z^{\left[ i \right]}_{\{F,V,A\}} + Feedforward(Z^{\left[ i \right]}_{\{F,V,A\}}))
\end{equation}

\subsection{Residual-based Cross-modal Fusion}
\label{sec: cs}

Our residual-based cross-modal fusion method could effectively fuse the information of three modalities with less information loss (on the left side of Figure \ref{fig:framework}). Specifically, the method accepts input for two modalities, which is called $\hat{X}_L$ and $F_{\{V,A\}}$. We obtain the mapping representations of the features for the two modalities by a linear projection. And then we process the two representations by $add$ and $tanh$ activation function. Finally, the fused representation $\hat{X}_L$ is obtained through $softmax$. We believe that the final fused information contains not only the complementary information of the three modalities, but also the features of the language modality:

\begin{equation}
    \hat{X}_F = softmax(tanh(L(\hat{X}_L) + L(F_{\{V,A\}}))) \in {R}^{T_l \times d_f}
\end{equation}
where $L$ stands for a linear projection.

In this process, in order to alleviate the information loss of language features, we use a residual connection between the fused representation and the original language representation. We use the algorithm \ref{alg:algorithm} to represent the entire process.

\section{Experiments}

\subsection{Datasets}

In this paper, we use three mainstream multimodal emotion recognition datasets: IEMOCAP, CMU-MOSI and CMU-MOSEI. The experiments are conducted on both the word-aligned and unaligned settings. The code will be publicly available after the paper is accepted.

\subsubsection{IEMOCAP.}

IEMOCAP \cite{iemocap} is a multimodal emotion recognition dataset that contains 151 videos along with corresponding transcripts and audios. In each video, two professional actors conduct dyadic conversations in English. Its intended data segmentation consists of 2,717 training samples, 798 validation samples and 938 test samples. The audio and visual features are extracted at the sampling frequencies of 12.5 Hz and 15 Hz, respectively. Although the human annotation has nine emotion categories, following the prior works \cite{2019work,dai2020modalitytransferable}, we take four categories: neutral, happy, sad, and angry. Moreover, this is a multi-label task (e.g., a person can feel sad and angry at the same time).  We report the binary classification accuracy and F1 scores for each emotion category according to \cite{Lv_2021_CVPR}.

\begin{table}[t]
    \centering
    \begin{tabular}{m{3cm}<{\centering} m{1.2cm}<{\centering} m{1cm}<{\centering} m{1.5cm}<{\centering}}
    \toprule
    Setting & CMU-MOSEI & CMU-MOSI & IEMOCAP \\ 
    \midrule
    Optimizer & Adam & Adam & Adam    \\
    Batch size   & 32 & 8 & 32    \\
    Learning rate   & 1e-3 & 2e-3 & 1e-3    \\
    Epochs & 120 & 100 & 60\\
    Feature size $d$ & 40 & 30 & 40    \\
    Attention head $h$ & 8 & 10 & 5    \\
    Kernel size (V/A) & 3/3 & 3/1 & 3/5    \\
    Transformer layer $D$ & 5 & 4 & 5    \\
    \bottomrule
    \end{tabular}
    \caption{ The hyperparameter settings adopted in each multimodal emotion recognition dataset.}
    \label{tab: hyper}
\end{table}

\begin{table}[t]
    \centering
    \begin{tabular}{m{1.6cm}<{\centering} m{1.5cm}<{\centering} m{1cm}<{\centering} m{1cm}<{\centering} m{1cm}<{\centering}}
    \toprule
    Method & \#Params(M) & $Acc_7(\%)$ & $Acc_2(\%)$ & $F1(\%)$ \\ 
    \midrule
    Conv1D & 0.38 & 50.6 & 78.5 & 80.1    \\
    \textbf{BiLSTM} & \textbf{0.41} & \textbf{51.8} & \textbf{80.9} & \textbf{81.5}     \\
    \hline
    $[$V, L$]$-\textgreater A & 0.41 & 50.7 & 79.2 & 80.8    \\
    $[$L, A$]$-\textgreater V & 0.41 & 51.1 & 79.3 & 81.0    \\
    \textbf{$[$V, A$]$-\textgreater L} & \textbf{0.41} & \textbf{51.8} & \textbf{80.9} & \textbf{81.5}    \\
    \bottomrule
    \end{tabular}
    \caption{Ablation study on the CMU-MOSEI dataset under the unaligned setting. [V,  A]-\textgreater L represents the integration of visual and audio modalities into language modalities to obtain the fused feature representation.}
    \label{tab: albation}
    \vspace{-5pt}
\end{table}

\begin{table*}[t]
    \centering
    \small
    \begin{tabular}{cccccccccc}
    \toprule
    \multirow{2}*{Setting} & \multirow{2}*{Method} & \multicolumn{2}{c}{Happy} & \multicolumn{2}{c}{Sad} & \multicolumn{2}{c}{Angry} & \multicolumn{2}{c}{Neutral} \\
    & & $Acc(\%)$ & $F1(\%)$ & $Acc(\%)$ & $F1(\%)$ & $Acc(\%)$ & $F1(\%)$ & $Acc(\%)$ & $F1(\%)$ \\ 
    \midrule
    \multirow{8}*{\textbf{Aligned}} & EF-LSTM & 86.0 & 84.2 & 80.2 & 80.5 & 85.2 & 84.5 & 67.8 & 67.1  \\
    & LF-LSTM & 85.1 & 86.3 & 78.9 & 81.7 & 84.7 & 83.0 & 67.1 & 67.6  \\
    & MFM &90.2 & 85.8 & 88.4 & 86.1 & 87.5 & 86.7 & 72.1 & 68.1  \\
    & RAVEN&87.3 & 85.8 & 83.4 & 83.1 & 87.3 & 86.7 & 69.7 & 69.3  \\
    & MCTN & 84.9 & 83.1 & 80.5 & 79.6 & 79.7 & 80.4 & 62.3 & 57.0  \\
    & MulT* & 86.4 & 82.9 & 82.3 & 82.4 & 85.3 & 85.8 & 71.2 & 70.0  \\
    & LMF-MulT & 85.3 & 84.1 & 84.1 & 83.4 & 85.7 & 86.2 & 71.2 & 70.8  \\
    & PMR† & 91.3 & 89.2 & 87.8 & 87.0 & 88.1 & 87.5 & 73.0 & 71.5  \\
    & \textbf{LMR-CBT(ours)} & \textbf{87.9} & \textbf{84.6} & \textbf{85.3} & \textbf{84.4} & \textbf{86.2} & \textbf{86.3} & \textbf{71.5} & \textbf{70.6}  \\
    \hline
    \multirow{8}*{\textbf{Unaligned}}& EF-LSTM  & 76.2 & 75.7 & 70.2 & 70.5 & 72.7 & 67.1 & 58.1 & 57.4  \\
    & LF-LSTM &72.5 & 71.8 & 72.9 & 70.4 & 68.6 & 67.9 & 59.6 & 56.2  \\
    & RAVEN &77.0 & 76.8 & 67.6 & 65.6 & 65.0 & 64.1 & 62.0 & 59.5  \\
    & MCTN &80.5 & 77.5 & 72.0 & 71.7 & 64.9 & 65.6 & 49.4 & 49.3  \\
    & MulT (1.07M)* & 85.6 & 79.0 & 79.4 & 70.3 & 75.8 & 65.4 & 59.5 & 44.7  \\
    & LMF-MulT (0.86M)  &85.6 & 79.0 & 79.4 & 70.3 & 75.8 & 65.4 & 59.2 & 44.0  \\
    & PMR (2.15M)†  & 86.4 & 83.3 & 78.5 & 75.3 & 75.0 & 71.3 & 63.7 & 60.9  \\
    & \textbf{LMR-CBT (0.34M)} & \textbf{85.7} & \textbf{79.5} & \textbf{79.4} & \textbf{72.6} & \textbf{76.0} & \textbf{70.7} & \textbf{63.6} & \textbf{60.5}    \\
    \bottomrule
    \end{tabular}
    \caption{ Comparison on the IEMOCAP dataset under both word-aligned setting and unaligned setting. The performance is evaluated by the binary classification accuracy and the F1 score for each emotion class. *: reproduced from open-source code; †: from \cite{Lv_2021_CVPR}. LMR-CBT achieves comparable and superior performance with only 0.34M parameters.}
    \label{tab: iemocap}
    \vspace{-5pt}
\end{table*}

\subsubsection{CMU-MOSI.}

CMU-MOSI \cite{cmumosi} is a dataset for multimodal emotion recognition and sentiment analysis, which comprises 2,199 short monologue video clips from 93 Youtube movie review videos. It contains 1,284 training samples, 229 validation samples and 686 test samples. The audio and visual features are extracted at the sampling frequencies of 12.5 Hz and 15 Hz, respectively. Human annotators label each sample with a sentiment score from -3 (strongly negative) to 3 (strongly positive). We use various metrics to evaluate the performance of the model, consistent with those used in previous work \cite{tsai2019multimodal}: 7-class accuracy (i.e. $Acc_7$), binary accuracy (i.e. $Acc_2$), and F1 score.

\subsubsection{CMU-MOSEI.}

CMU-MOSEI \cite{cmu-mosei} is also a dataset for multimodal emotion recognition and sentiment analysis, which contains 3,837 videos from 1,000 diverse speakers. Its pre-determined data segmentation includes 16,326 training samples, 1,871 validation samples and 4,659 test samples. The audio and visual features are extracted at the sampling frequencies of 20 Hz and 15 Hz, respectively. In addition, each data sample is also annotated with a sentiment scores on a Likert scale [-3, 3]. We use the same performance metrics as above.

\subsection{Implementation details}

For feature extraction of the language modality, we convert video transcripts into pre-trained Glove \cite{glove} model to obtain 300-dimensional word embeddings. For feature extraction of visual modality, we use Facet \cite{facenet} to represent 35 facial action units, which record facial muscle movements for representing basic and high-level emotions in each frame. For the audio modality, we use COVAREP \cite{covarep} for extracting acoustic signals to obtain 74-dimensional vectors.

Table 1 shows the hyperparameters used in training and testing for each dataset. The kernel size is used to process the input sequences for the audio and visual modalities, and since BiLSTM is used for the language modality, no kernel size is involved. We train our model on a single RTX 2080Ti. The details are provided in the supplementary file.

\subsection{Comparison with the State-of-the-arts}

We compare the proposed approach with the existing state-of-the-art methods, including Early Fusion LSTM (EF-LSTM), Late Fusion LSTM (LF-LSTM), Multimodal Factorization Model (MFM) \cite{tsai2019learning}, Graph-MFN (GMFN), Recurrent Attended Variation Embedding Network (RAVEN) \cite{2019work}, Multimodal Cyclic Translation Network (MCTN) \cite{mctn}, Multimodal Transformer (MulT) \cite{tsai2019multimodal}, Low Rank Fusion based Transformers (LMF-MulT) \cite{sahay2020low}, Modality-Invariant and -Specific Representations (MISA) \cite{hazarika2020misa}, Progressive Modality Reinforcement (PMR) \cite{Lv_2021_CVPR}. Among these methods, LF-LSTM, MulT, LMF-MulT, and PMR can be directly applied the unaligned setting. For the other methods, we introduce the connectionist temporal classification (CTC) \cite{10.1145/1143844.1143891} module to make them applicable to unaligned settings.

\subsubsection{Word-aligned setting.}

This setting requires manual alignment of language words with visual and audio. We show the comparison of our method with other benchmarks in the upper part of Tables \ref{tab: iemocap}-\ref{tab: mosei}. The experimental results show that the proposed method achieves a comparable performance level to PMR \cite{Lv_2021_CVPR} on different metrics for the three datasets. Compared with LMF-MulT~\cite{sahay2020low}, which uses six transformer encoders, we achieve better performance on different datasets using half of the transformer encoders.

\subsubsection{Unaligned setting.}

This setting is more challenging than the word-aligned setting, where cross-modal information is extracted directly from unaligned multimodal sequences to classify emotions. We show the comparison of our approach with other benchmarks in the lower part of Tables \ref{tab: iemocap}-\ref{tab: mosei}. Moreover, Figure \ref{fig:params} demonstrates that our proposed model reaches the state-of-the-art with a minimum number of parameters (only 0.41M) on the CMU-MOSEI dataset. Compared with other approaches, our proposed light-weight network is more applicable to real scenarios. We can draw the following conclusions from the experimental results:
\begin{itemize}
    \item With the exception of MulT \cite{tsai2019multimodal}, LMF-MulT \cite{sahay2020low}, and PMR \cite{Lv_2021_CVPR}, most of the models perform poorly in the unaligned setting because they do not take into account the interactions between the modalities. In addition, the outstanding performance of MISA \cite{hazarika2020misa} is due to the pre-trained model, which contains a large number of parameters.
    \item Compared to LMF-MulT and MulT models, our approach outperforms in different metrics. Compared to PMR, we have comparable or better performance on different datasets with a minimal number of parameters.
    \item What's more, the number of parameters of MISA and PMR on the CMU-MOSEI dataset reaches 15.9 M and 2.15 M, respectively, while our proposed method uses only 0.41 M. For MISA, the number of parameters is equivalent to 38 times that of our proposed method, while PMR is equivalent to as much as 6 times.
\end{itemize}

\begin{table}[t]
    \centering \small
    \begin{tabular}{m{1.0cm}<{\centering} m{2.7cm}<{\centering} m{0.9cm}<{\centering} m{0.9cm}<{\centering} m{0.9cm}<{\centering}}
    \toprule
    Setting & Method & $Acc_7(\%)$ & $Acc_2(\%)$ & $F1(\%)$ \\ 
    \midrule
    \multirow{8}*{\textbf{Aligned}} & EF-LSTM & 33.7 & 75.3 & 75.2    \\
    & LF-LSTM & 35.3 & 76.8 & 76.7    \\
    & MFM & 36.2 & 78.1 & 78.1    \\
    & RAVEN & 33.2 & 78.0 & 76.6    \\
    & MCTN & 35.6 & 79.3 & 79.1    \\
    & MulT* & 33.1 & 78.5 & 78.4    \\
    & LMF-MulT & 32.4 & 77.9 & 77.9   \\
    & PMR† & 40.6 & 83.6 & 83.4    \\
    & \textbf{LMR-CBT(ours)} & \textbf{39.2} & \textbf{81.6} & \textbf{79.8}    \\
    \hline
    \multirow{8}*{\textbf{Unaligned}}& EF-LSTM & 31.0 & 73.6 & 74.5    \\
    & LF-LSTM & 33.7 & 77.6 & 77.8    \\
    & RAVEN & 31.7 & 72.7 & 73.1    \\
    & MCTN & 32.7 & 75.9 & 76.4    \\
    & MulT (1.07M)* & 34.3 & 80.3 & 80.4    \\
    & LMF-MulT (0.84M) & 34.0 & 78.5 & 78.5   \\
    & MISA (15.9M)‡ & 41.4 & 81.8 & 81.8 \\
    & PMR (2.14M)† & 40.6 & 82.4 & 82.1    \\
    & \textbf{LMR-CBT (0.35M)} & \textbf{39.5} & \textbf{81.2} & \textbf{81.0}    \\
    \bottomrule
    \end{tabular}
    \caption{Comparison on the CMU-MOSI dataset under both word-aligned setting and unaligned setting. *: reproduced from open-source code; †: from \cite{Lv_2021_CVPR}; ‡: from \cite{yu2021learning}. LMR-CBT achieves comparable and superior performance with only 0.35M parameters. }
    \label{tab: mosi}
    \vspace{-5pt}
\end{table}

\begin{table}[t]
    \centering \small
    \begin{tabular}{m{1.0cm}<{\centering} m{2.7cm}<{\centering} m{0.9cm}<{\centering} m{0.9cm}<{\centering} m{0.9cm}<{\centering}}
    \toprule
    Setting & Method & $Acc_7(\%)$ & $Acc_2(\%)$ & $F1(\%)$ \\ 
    \midrule
    \multirow{8}*{\textbf{Aligned}} & EF-LSTM & 47.4 & 78.2 & 77.9    \\
    & LF-LSTM & 48.8 & 80.6 & 80.6    \\
    & G-MFN & 45.0 & 76.9 & 77.0    \\
    & RAVEN & 50.0 & 79.1 & 79.5    \\
    & MCTN & 49.6 & 79.8 & 80.6    \\
    & MulT* & 49.3 & 80.5 & 81.1    \\
    & LMF-MulT & 50.2 & 80.3 & 80.3   \\
    & PMR† & 52.5 & 83.3 & 82.6    \\
    & \textbf{LMR-CBT(ours)} & \textbf{50.7} & \textbf{80.5} & \textbf{80.9}    \\
    \hline
    \multirow{8}*{\textbf{Unaligned}}& EF-LSTM & 46.3 & 76.1 & 75.9    \\
    & LF-LSTM & 48.8 & 77.5 & 78.2    \\
    & RAVEN & 45.5 & 75.4 & 75.7    \\
    & MCTN & 48.2 & 79.3 & 79.7    \\
    & MulT (1.07M)* & 50.4 & 80.7 & 80.6    \\
    & LMF-MulT (0.86M) & 49.3 & 80.8 & 81.3   \\
    & MISA (15.9M)‡ & 52.1 & 80.7 & 81.1 \\
    & PMR (2.15M)† & 51.8 & 83.1 & 82.8   \\
    & \textbf{LMR-CBT (0.41M)} & \textbf{51.8} & \textbf{80.9} & \textbf{81.5}    \\
    \bottomrule
    \end{tabular}
    \caption{Comparison on the CMU-MOSEI dataset under both word-aligned setting and unaligned setting. *: reproduced from open-source code; †: from \cite{Lv_2021_CVPR}; ‡: from \cite{yu2021learning}. LMR-CBT achieves comparable and superior performance with only 0.41M parameters.}
    \label{tab: mosei}
    \vspace{-5pt}
\end{table}

\subsection{Ablation study}

\subsubsection{Effectiveness of BiLSTM.}

For the language modality, we adopt BiLSTM to capture the long-time dependency and the contextual information association between texts. We replace BiLSTM with Conv1D for the comparison of the experiments, and the experimental results (on the upper part of Table \ref{tab: albation}) demonstrate that compared to Conv1D, there is a dramatic noticeable improvement in performance despite a marginal increase in the number of parameters, with a 1.4\% higher F1 score. This indicates that BiLSTM is more suitable for processing textual information and can adequately represent the features of the linguistic modality.

\subsubsection{Effectiveness of CB-Transformer.}

To implement a efficient  cross-modal fusion mechanism, we integrate deep audio/visual features with the shallow language features, and this could be denoted by [V, A]-\textgreater L. We compare the different operations of the three modalities in feature fusion. Specifically, [V,  L]-\textgreater A denotes the integration of visual and speech modalities into audio modalities to obtain fused features, and [L, A]-\textgreater V denotes the integration of speech and audio modalities into visual modalities to obtain fused features. From the experimental results, as shown in the lower part of Table \ref{tab: albation}, [V, A]-\textgreater L achieves the best performance compared to the remaining two settings with the same number of parameters. Meanwhile, we note that the results are the worst when we obtain the fused features through the audio, which indicates that we do not obtain a high-level feature representation of the audio. Moreover, we analyze the reason is that the BiLSTM already has a good representation of the language modality in the feature processing stage and can make the performance work out.

\section{Conclusion and Future Work}

In this paper, we propose a neural network to learn modality-fused representations with CB-Transformer (LMR-CBT) for multimodal emotion recognition from unaligned multimodal sequences. First of all, we perform feature preprocessing on each modality respectively. Unlike previous work, we use BiLSTM for the language modality to handle long-term dependencies and contextual information. Furthermore, we design a novel transformer with cross-modal blocks (CB-Transformer) that enables complementary learning of different modalities, which is mainly divided into local temporal learning, cross modal feature fusion and global self-attention representations. The CB-Transformer can represent the fused features without losing the original features, and can process unaligned multimodal sequences efficiently. Finally, we apply the proposed method to IEMOCAP, CMU-MOSI and CMU-MOSEI, respectively, and the experimental results show that our proposed method achieves comparable or better results compared to the existing state-of-the-art methods with the minimum number of parameters.

We also find that the initial features of the three modalities are highly important but limited by preprocessing. In future work, we will build an end-to-end multimodal learning network and introduce the learning of more modalities, such as body postures, to explore the relationship between different modalities.

\bibliography{aaai22}

\end{document}